# An adaptive algorithm for finite stochastic partial monitoring

**Gábor Bartók**                                    BARTOK@UALBERTA.CA
**Navid Zolghadr**                                  ZOLGHADR@UALBERTA.CA
**Csaba Szepesvári**                                SZEPESVA@UALBERTA.CA
Department of Computing Science, University of Alberta, AB, Canada, T6G 2E8

## Abstract

We present a new anytime algorithm that achieves near-optimal regret for any instance of finite stochastic partial monitoring. In particular, the new algorithm achieves the minimax regret, within logarithmic factors, for both "easy" and "hard" problems. For easy problems, it additionally achieves logarithmic individual regret. Most importantly, the algorithm is adaptive in the sense that if the opponent strategy is in an "easy region" of the strategy space then the regret grows as if the problem was easy. As an implication, we show that under some reasonable additional assumptions, the algorithm enjoys an $O(\sqrt{T})$ regret in Dynamic Pricing, proven to be hard by Bartók et al. (2011).

## 1. Introduction

Partial monitoring can be cast as a sequential game played by a learner and an opponent. In every time step, the learner chooses an action and simultaneously the opponent chooses an outcome. Then, based on the action and the outcome, the learner suffers some loss and receives some feedback. Neither the outcome nor the loss are revealed to the learner. A partial-monitoring game with $N$ actions and $M$ outcomes is defined with the pair $G = (L, H)$, where $L \in \mathbb{R}^{N \times M}$ is the *loss matrix*, and $H \in \Sigma^{N \times M}$ is the *feedback matrix* over some arbitrary set of symbols $\Sigma$. These matrices are announced to both the learner and the opponent before the game starts. At time step $t$, if $I_t \in \underline{N} = \{1, 2, \ldots, N\}$ and $J_t \in \underline{M} = \{1, 2, \ldots, M\}$ denote the (possibly random) choices of the learner and the opponent, respectively then, the loss suffered by the learner in that time step is $L[I_t, J_t]$, while the feedback received is $H[I_t, J_t]$.

The goal of the learner (or player) is to minimize his cumulative loss $\sum_{t=1}^{T} L[I_t, J_t]$. The performance of the learner is measured in terms of the *regret*, defined as the excess cumulative loss he suffers compared to that of the best fixed action in hindsight:

$$R_T = \sum_{t=1}^{T} L[I_t, J_t] - \min_{i \in \underline{N}} \sum_{t=1}^{T} L[i, J_t].$$

The regret usually grows with the time horizon $T$. What distinguishes between a "successful" and an "unsuccessful" learner is the growth rate of the regret. A regret linear in $T$ means that the learner does not approach the performance of the optimal action. On the other hand, if the growth rate is sublinear, it is said that the learner can learn the game.

In this paper we restrict our attention to *stochastic* games, adding the extra assumption that the opponent generates the outcomes with a sequence of independent and identically distributed random variables. This distribution will be called the *opponent strategy*. As for the player, a *player strategy* (or algorithm) is a (possibly random) function from the set of feedback sequences (observation histories) to the set of actions.

In stochastic games, we use a slightly different notion of regret: we compare the cumulative loss with that of the action with the lowest expected loss.

$$R_T = \sum_{t=1}^{T} L[I_t, J_t] - T \min_{i \in \underline{N}} \mathbb{E}[L[i, J_1]].$$

The "hardness" of a game is defined in terms of the *minimax expected regret* (or minimax regret for short):

$$R_T(G) = \min_{\mathcal{A}} \max_{p \in \Delta_M} \mathbb{E}[R_T],$$

where $\Delta_M$ is the space of opponent strategies, and $\mathcal{A}$ is any strategy of the player. In other words, the minimax regret is the worst-case expected regret of the best algorithm.

A question of major importance is how the minimax regret scales with the parameters of the game, such as the time horizon $T$, the number of actions $N$, the number of outcomes $M$. In the stochastic setting, another measure of "hardness" is worth studying, namely the *individual* or *problem-dependent* regret, defined as the expected regret given a fixed opponent strategy.

### 1.1. Related work

Two special cases of partial monitoring have been extensively studied for a long time: *full-information* games, where the feedback carries enough information for the learner to infer the outcome for any action-outcome pair, and *bandit* games, where the learner receives the loss of the chosen action as feedback. Since Vovk (1990) and Littlestone & Warmuth (1994) we know that for full-information games, the minimax regret scales as $\Theta(\sqrt{T \log N})$. For bandit games, the minimax regret has been proven to scale as $\Theta(\sqrt{NT})$ (Audibert & Bubeck, 2009).[1] The individual regret of these kind of games has also been studied: Auer et al. (2002) showed that given any opponent strategy, the expected regret can be upper bounded by $c \sum_{i \in \underline{N}: \delta_i \neq 0} \frac{1}{\delta_i} \log T$, where $\delta_i$ is the expected difference between the loss of action $i$ and an optimal action.

Finite partial monitoring problems were introduced by Piccolboni & Schindelhauer (2001). They proved that a game is either "hopeless" (that is, its minimax regret scales linearly with $T$), or the regret can be upper bounded by $O(T^{3/4})$. They also give a characterization of hopeless games. Namely, a game is hopeless if it does not satisfy the *global observability* condition (see Definition 5 in Section 2). Their upper bound for non-hopeless games was tightened to $O(T^{2/3})$ by Cesa-Bianchi et al. (2006), who also showed that there exists a game with a matching lower bound.

Cesa-Bianchi et al. (2006) posted the problem of characterizing partial-monitoring games with minimax regret less than $\Theta(T^{2/3})$. This problem has been solved since then. The first steps towards classifying partial-monitoring games were made by Bartók et al. (2010), who characterized almost all games with *two outcomes*. They proved that there are only four categories: games with minimax regret $0$, $\widetilde{\Theta}(\sqrt{T})$, $\Theta(T^{2/3})$, and $\Theta(T)$, and named them *trivial*, *easy*, *hard*, and *hopeless*, re-spectively.[2] They also found that there exist games that are easy, but can not easily be "transformed" to a bandit or full-information game. Later, Bartók et al. (2011) proved the same results for finite stochastic partial monitoring, with any finite number of outcomes. The condition that separates easy games from hard games is the *local observability* condition (see Definition 6). The algorithm BALATON introduced there works by eliminating actions that are thought to be suboptimal with high confidence. They conjectured in their paper that the same classification holds for non-stochastic games, without changing the condition. Recently, Foster & Rakhlin (2011) designed the algorithm NEIGHBORHOODWATCH that proves this conjecture to be true. Foster & Rakhlin prove an upper bound on a stronger notion of regret, called *internal regret*.

### 1.2. Contributions

In this paper, we extend the results of Bartók et al. (2011). We introduce a new algorithm, called CBP for "**C**onfidence **B**ound **P**artial monitoring", with various desirable properties. First of all, while BALATON only works on easy games, CBP can be run on any non-hopeless game, and it achieves (up to logarithmic factors) the minimax regret rates both for easy and hard games (see Corollaries 3 and 2). Furthermore, it also achieves logarithmic problem-dependent regret for easy games (see Corollary 1). It is also an "anytime" algorithm, meaning that it does not have to know the time horizon, nor does it have to use the doubling trick, to achieve the desired performance.

The final, and potentially most impactful, aspect of our algorithm is that through additional assumptions on the set of opponent strategies, the minimax regret of even hard games can be brought down to $\widetilde{\Theta}(\sqrt{T})$! While this statement may seem to contradict the result of Bartók et al. (2011), in fact it does not. For the precise statement, see Theorem 2. We call this property "adaptiveness" to emphasize that the algorithm does not even have to know that the set of opponent strategies is restricted.

## 2. Definitions and notations

Recall from the introduction that an instance of partial monitoring with $N$ actions and $M$ outcomes is defined by the pair of matrices $L \in \mathbb{R}^{N \times M}$ and $H \in \Sigma^{N \times M}$, where $\Sigma$ is an arbitrary set of symbols. In each round $t$, the opponent chooses an outcome $J_t \in \underline{M}$ and simultaneously the learner chooses an action $I_t \in \underline{N}$. Then, the feedback $H[I_t, J_t]$ is revealed and the learner suffers the loss $L[I_t, J_t]$. It is important to note that the loss is not revealed to the learner.

---

[1] The Exp3 algorithm due to Auer et al. (2003) achieves almost the same regret, with an extra logarithmic term.

[2] Note that these results do not concern the growth rate in terms of other parameters (like $N$).

As it was previously mentioned, in this paper we deal with stochastic opponents only. In this case, the choice of the opponent is governed by a sequence $J_1, J_2, \ldots$ of i.i.d. random variables. The distribution of these variables $p \in \Delta_M$ is called an *opponent strategy*, where $\Delta_M$, also called the *probability simplex*, is the set of all distributions over the $M$ outcomes. It is easy to see that, given opponent strategy $p$, the expected loss of action $i$ can be expressed as $\ell_i^\top p$, where $\ell_i$ is defined as the column vector consisting of the $i^{\text{th}}$ row of $L$.

The following definitions, taken from Bartók et al. (2011), are essential for understanding how the structure of $L$ and $H$ determines the "hardness" of a game.

Action $i$ is called *optimal* under strategy $p$ if its expected loss is not greater than that of any other action $i' \in \underline{N}$. That is, $\ell_i^\top p \le \ell_{i'}^\top p$. Determining which action is optimal under opponent strategies yields the *cell decomposition*[3] of the probability simplex $\Delta_M$:

**Definition 1** (Cell decomposition). *For every action $i \in \underline{N}$, let $\mathcal{C}_i = \{p \in \Delta_M : \text{action } i \text{ is optimal under } p\}$. The sets $\mathcal{C}_1, \ldots, \mathcal{C}_N$ constitute the cell decomposition of $\Delta_M$.*

Now we can define the following important properties of actions:

**Definition 2** (Properties of actions).
- *Action $i$ is called* dominated *if $\mathcal{C}_i = \emptyset$. If an action is not dominated then it is called* non-dominated.

- *Action $i$ is called* degenerate *if it is non-dominated and there exists an action $i'$ such that $\mathcal{C}_i \subsetneq \mathcal{C}_{i'}$.*

- *If an action is neither dominated nor degenerate then it is called* Pareto-optimal. *The set of Pareto-optimal actions is denoted by $\mathcal{P}$.*

From the definition of cells we see that a cell is either empty or it is a closed polytope. Furthermore, Pareto-optimal actions have $(M - 1)$-dimensional cells. The following definition, important for our algorithm, also uses the dimensionality of polytopes:

**Definition 3** (Neighbors). *Two Pareto-optimal actions $i$ and $j$ are* neighbors *if $\mathcal{C}_i \cap \mathcal{C}_j$ is an $(M - 2)$-dimensional polytope. Let $\mathcal{N}$ be the set of unordered pairs over $\underline{N}$ that contains neighboring action-pairs. The* neighborhood action set *of two neighboring actions $i$, $j$ is defined as $N_{i,j}^+ = \{k \in \underline{N} : \mathcal{C}_i \cap \mathcal{C}_j \subseteq \mathcal{C}_k\}$.*

---

[3]The concept of cell decomposition also appears in Piccolboni & Schindelhauer (2001).

Note that the neighborhood action set $N_{i,j}^+$ naturally contains $i$ and $j$. If $N_{i,j}^+$ contains some other action $k$ then either $\mathcal{C}_k = \mathcal{C}_i$, $\mathcal{C}_k = \mathcal{C}_j$, or $\mathcal{C}_k = \mathcal{C}_i \cap \mathcal{C}_j$.

In general, the elements of the feedback matrix $H$ can be arbitrary symbols. Nevertheless, the nature of the symbols themselves does not matter in terms of the structure of the game. What determines the feedback structure of a game is the occurrence of identical symbols in each row of $H$. To "standardize" the feedback structure, the *signal matrix* is defined for each action:

**Definition 4.** *Let $s_i$ be the number of distinct symbols in the $i^{\text{th}}$ row of $H$ and let $\sigma_1, \ldots, \sigma_{s_i} \in \Sigma$ be an enumeration of those symbols. Then the* signal matrix $S_i \in \{0, 1\}^{s_i \times M}$ *of action $i$ is defined as $S_i[k, l] = \mathbb{I}_{\{H[i,l] = \sigma_k\}}$.*

The idea of this definition is that if $p \in \Delta_M$ is the opponent's strategy then $S_i p$ gives the distribution over the symbols underlying action $i$. In fact, it is also true that observing $H[I_t, J_t]$ is equivalent to observing the vector $S_{I_t} e_{J_t}$, where $e_k$ is the $k^{\text{th}}$ unit vector in the standard basis of $\mathbb{R}^M$. From now on we assume without loss of generality that the learner's observation at time step $t$ is the random vector $Y_t = S_{I_t} e_{J_t}$. Note that the dimensionality of this vector depends on the action chosen by the learner, namely $Y_t \in \mathbb{R}^{s_{I_t}}$.

The following two definitions play a key role in classifying partial-monitoring games based on their difficulty.

**Definition 5** (Global observability (Piccolboni & Schindelhauer, 2001)). *A partial-monitoring game $(L, H)$ admits the* global observability *condition, if for all pairs $i, j$ of actions, $\ell_i - \ell_j \in \oplus_{k \in \underline{N}} \operatorname{Im} S_k^\top$.*

**Definition 6** (Local observability (Bartók et al., 2011)). *A pair of neighboring actions $i, j$ is said to be* locally observable *if $\ell_i - \ell_j \in \oplus_{k \in N_{i,j}^+} \operatorname{Im} S_k^\top$. We denote by $\mathcal{L} \subseteq \mathcal{N}$ the set of locally observable pairs of actions (the pairs are unordered). A game satisfies the* local observability *condition if every pair of neighboring actions is locally observable, i.e., if $\mathcal{L} = \mathcal{N}$.*

The main result of Bartók et al. (2011) is that locally observable games have $\widetilde{O}(\sqrt{T})$ minimax regret. It is easy to see that local observability implies global observability. Also, from Piccolboni & Schindelhauer (2001) we know that if global observability does not hold then the game has linear minimax regret. From now on, we only deal with games that admit the global observability condition.

A collection of the concepts and symbols introduced in this section is shown in Table 1.

Table 1. List of basic symbols

| Symbol | Definition | Found in/at |
|---|---|---|
| $N, M \in \mathbb{N}$ | number of actions and outcomes | |
| $\underline{N}$ | $\{1, \ldots, N\}$, set of actions | |
| $\Delta_M \subset \mathbb{R}^M$ | $M$-dim. simplex, set of opponent strategies | |
| $p^* \in \Delta_M$ | opponent strategy | |
| $L \in \mathbb{R}^{N \times M}$ | loss matrix | |
| $H \in \Sigma^{N \times M}$ | feedback matrix | |
| $\ell_i \in \mathbb{R}^M$ | $\ell_i = L[i, :]$, loss vector underlying action $i$ | |
| $\mathcal{C}_i \subseteq \Delta_M$ | cell of action $i$ | Definition 1 |
| $\mathcal{P} \subseteq \underline{N}$ | set of Pareto-optimal actions | Definition 2 |
| $\mathcal{N} \subseteq \underline{N}^2$ | set of unordered neighboring action-pairs | Definition 3 |
| $N_{i,j}^+ \subseteq \underline{N}$ | neighborhood action set of $\{i, j\} \in \mathcal{N}$ | Definition 3 |
| $S_i \in \{0, 1\}^{s_i \times M}$ | signal matrix of action $i$ | Definition 4 |
| $\mathcal{L} \subseteq \mathcal{N}$ | set of locally observable action pairs | Definition 6 |
| $V_{i,j} \subseteq \underline{N}$ | observer actions underlying $\{i, j\} \in \mathcal{N}$ | Definition 7 |
| $v_{i,j,k} \in \mathbb{R}^{s_k}, \, k \in V_{i,j}$ | observer vectors | Definition 7 |
| $W_i \in \mathbb{R}$ | confidence width for action $i \in \underline{N}$ | Definition 7 |

## 3. The proposed algorithm

Our algorithm builds on the core idea underlying algorithm BALATON of Bartók et al. (2011), so we start with a brief review of BALATON. BALATON uses sweeps to successively eliminate suboptimal actions. This is done by estimating the differences between the expected losses of pairs of actions, *i.e.*, $\delta_{i,j} = (\ell_i - \ell_j)^\top p^*$ $(i, j \in \underline{N})$. In fact, BALATON exploits that it suffices to keep track of $\delta_{i,j}$ for *neighboring pairs* of actions (*i.e.*, for action pairs $i, j$ such that $\{i, j\} \in \mathcal{N}$). This is because if an action $i$ is suboptimal, it will have a neighbor $j$ that has a smaller expected loss and so the action $i$ will get eliminated when $\delta_{i,j}$ is checked. Now, to estimate $\delta_{i,j}$ for some $\{i, j\} \in \mathcal{N}$ one observes that under the local observability condition, it holds that $\ell_i - \ell_j = \sum_{k \in N_{i,j}^+} S_k^\top v_{i,j,k}$ for some vectors $v_{i,j,k} \in \mathbb{R}^{\sigma_k}$. This yields that $\delta_{i,j} = (\ell_i - \ell_j)^\top p^* = \sum_{k \in N_{i,j}^+} v_{i,j,k}^\top S_k p^*$. Since $\nu_k \stackrel{\text{def}}{=} S_k p^*$ is the vector of the distribution of symbols under action $k$, which can be estimated by $\nu_k(t)$, the empirical frequencies of the individual symbols observed under $k$ up to time $t$, BALATON uses $\sum_{k \in N_{i,j}^+} v_{i,j,k}^\top \nu_k(t)$ to estimate $\delta_{i,j}$. Since none of the actions in $N_{i,j}^+$ can get eliminated before one of $\{i, j\}$ gets eliminated, the estimate of $\delta_{i,j}$ gets refined until one of $\{i, j\}$ is eliminated.

The essence of why BALATON achieves a low regret is as follows: When $i$ is not a neighbor of the optimal action $i^*$ one can show that it will be eliminated before all neighbors $j$ "between $i$ and $i^*$" get eliminated. Thus, the contribution of such "far" actions to the regret is minimal. When $i$ is a neighbor of $i^*$, it will be eliminated in time proportional to $\delta_{i,i^*}^{-2}$. Thus the contribution to the regret of such an action is proportional to $\delta_i^{-1}$, where $\delta_i \stackrel{\text{def}}{=} \delta_{i,i^*}$. It also holds that the contribution to the regret of $i$ cannot be larger than $\delta_i T$. Thus, the contribution of $i$ to the regret is at most $\min(\delta_i T, \delta_i^{-1}) \leq \sqrt{T}$.

When some pairs $\{i, j\} \in \mathcal{N}$ are not locally observable, one needs to use actions other than those in $N_{i,j}^+$ to construct an estimate of $\delta_{i,j}$. Under global observability, $\ell_i - \ell_j = \sum_{k \in V_{i,j}} S_k^\top v_{i,j,k}$ for an appropriate subset $V_{i,j} \subset \underline{N}$ and an appropriate set of vectors $v_{i,j,\cdot}$. Thus, if the actions in $V_{i,j}$ are kept in play, one can estimate the difference $\delta_{i,j}$ as before, using $\sum_{k \in V_{i,j}} v_{i,j,k}^\top \nu_k(t)$. This motivates the following definition:

**Definition 7** (Observer sets and observer vectors). *The observer set $V_{i,j} \subset \underline{N}$ underlying a pair of neighboring actions $\{i, j\} \in \mathcal{N}$ is a set of actions such that*

$$\ell_i - \ell_j \in \oplus_{k \in V_{i,j}} \operatorname{Im} S_k^\top .$$

*The observer vectors $(v_{i,j,k})_{k \in V_{i,j}}$ are defined to satisfy the equation $\ell_i - \ell_j = \sum_{k \in V_{i,j}} S_k^\top v_{i,j,k}$. In particular, $v_{i,j,k} \in \mathbb{R}^{s_k}$. In what follows, the choice of the observer sets and vectors is restricted so that $V_{i,j} = V_{j,i}$ and $v_{i,j,k} = -v_{j,i,k}$. Furthermore, the observer set $V_{i,j}$ is constrained to be a superset of $N_{i,j}^+$ and in particular when a pair $\{i, j\}$ is locally observable, $V_{i,j} = N_{i,j}^+$ must hold. Finally, for any action $k \in \bigcup_{\{i,j\} \in \mathcal{N}} N_{i,j}^+$, let $W_k = \max_{i,j:k \in V_{i,j}} \|v_{i,j,k}\|_\infty$ be the* confidence width *of action $k$.*

The reason of the particular choice $V_{i,j} = N_{i,j}^+$ for lo-cally observable pairs $\{i, j\}$ is that we plan to use $V_{i,j}$ (and the vectors $v_{i,j,\cdot}$) in the case of locally observable pairs, too. For not locally observable pairs, the whole action set $\underline{N}$ is always a valid observer set (thus, $V_{i,j}$ can be found). However, whenever possible, it is better to use a smaller set. The actual choice of $V_{i,j}$ (and $v_{i,j,k}$) is postponed until the effect of this choice on the regret becomes clear.

With the observer sets, the basic idea of the algorithm becomes as follows: (i) Eliminate the suboptimal actions in successive sweeps; (ii) In each sweep, enrich the set of remaining actions $\mathcal{P}(t)$ by adding the observer actions underlying the remaining neighboring pairs $\{i, j\} \in \mathcal{N}(t)$: $\mathcal{V}(t) = \bigcup_{\{i,j\} \in \mathcal{N}(t)} V_{i,j}$; (iii) Explore the actions in $\mathcal{P}(t) \cup \mathcal{V}(t)$ to update the symbol frequency estimate vectors $\nu_k(t)$. Another refinement is to eliminate the sweeps so as to make the algorithm enjoy an advantageous anytime property. This can be achieved by selecting in each step only one action. We propose the action to be chosen should be the one that maximizes the reduction of the remaining uncertainty.

This algorithm could be shown to enjoy $\sqrt{T}$ regret for locally observable games. However, if we run it on a non-locally observable game and the opponent strategy is on $\mathcal{C}_i \cap \mathcal{C}_j$ for $\{i, j\} \in \mathcal{N} \setminus \mathcal{L}$, it will suffer linear regret! The reason is that if both actions $i$ and $j$ are optimal, and thus never get eliminated, the algorithm will choose actions from $V_{i,j} \setminus N_{i,j}^+$ too often. Furthermore, even if the opponent strategy is not on the boundary the regret can be too high: say action $i$ is optimal but $\delta_j$ is small, while $\{i, j\} \in \mathcal{N} \setminus \mathcal{L}$. Then a third action $k \in V_{i,j}$ with large $\delta_k$ will be chosen proportional to $1/\delta_j^2$ times, causing high regret. To combat this we restrict the frequency with which an action can be used for "information seeking purposes". For this, we introduce the set of rarely chosen actions,

$$\mathcal{R}(t) = \{k \in \underline{N} : n_k(t) \le \eta_k f(t)\},$$

where $\eta_k \in \mathbb{R}$, $f : \mathbb{N} \to \mathbb{R}$ are tuning parameters to be chosen later. Then, the set of actions available at time $t$ is restricted to $\mathcal{P}(t) \cup N^+(t) \cup (\mathcal{V}(t) \cap \mathcal{R}(t))$, where $N^+(t) = \bigcup_{\{i,j\} \in \mathcal{N}(t)} N_{i,j}^+$. We will show that with these modifications, the algorithm achieves $O(T^{2/3})$ regret in the general case, while it will also be shown to achieve an $O(\sqrt{T})$ regret when the opponent uses a benign strategy. A pseudocode for the algorithm is given in Algorithm 1.

It remains to specify the function GETPOLYTOPE. It gets the array *halfSpace* as input. The array *halfSpace* stores which neighboring action pairs have a confident estimate on the difference of their expected losses, along with the sign of the difference (if confi-

---

**Algorithm 1** CBP

**Input:** $L$, $H$, $\alpha$, $\eta_1, \ldots, \eta_N$, $f = f(\cdot)$
Calculate $\mathcal{P}$, $\mathcal{N}$, $V_{i,j}$, $v_{i,j,k}$, $W_k$
**for** $t = 1$ **to** $N$ **do**
    Choose $I_t = t$ and observe $Y_t$     {Initialization}
    $n_{I_t} \leftarrow 1$     {# times the action is chosen}
    $\nu_{I_t} \leftarrow Y_t$     {Cumulative observations}
**end for**
**for** $t = N + 1, N + 2, \ldots$ **do**
    **for each** $\{i, j\} \in \mathcal{N}$ **do**
        $\tilde{\delta}_{i,j} \leftarrow \sum_{k \in V_{i,j}} v_{i,j,k}^\top \frac{\nu_k}{n_k}$   {Loss diff. estimate}
        $c_{i,j} \leftarrow \sum_{k \in V_{i,j}} \|v_{i,j,k}\|_\infty \sqrt{\frac{\alpha \log t}{n_k}}$   {Confidence}
        **if** $|\tilde{\delta}_{i,j}| \ge c_{i,j}$ **then**
            $halfSpace(i, j) \leftarrow \mathrm{sgn}\,\tilde{\delta}_{i,j}$
        **else**
            $halfSpace(i, j) \leftarrow 0$
        **end if**
    **end for**
    $[\mathcal{P}(t), \mathcal{N}(t)] \leftarrow$ GETPOLYTOPE$(\mathcal{P}, \mathcal{N}, halfSpace)$
    $N^+(t) = \cup_{\{i,j\} \in \mathcal{N}(t)} N_{ij}^+$
    $\mathcal{V}(t) = \cup_{\{i,j\} \in \mathcal{N}(t)} V_{ij}$
    $\mathcal{R}(t) = \{k \in \underline{N} : n_k(t) \le \eta_k f(t)\}$
    $\mathcal{S}(t) = \mathcal{P}(t) \cup N^+(t) \cup (\mathcal{V}(t) \cap \mathcal{R}(t))$
    Choose $I_t = \operatorname{argmax}_{i \in \mathcal{S}(t)} \frac{W_i^2}{n_i}$ and observe $Y_t$
    $\nu_{I_t} \leftarrow \nu_{I_t} + Y_t$
    $n_{I_t} \leftarrow n_{I_t} + 1$
**end for**

---

dent). Each of these confident pairs define an open halfspace, namely

$$\Delta_{\{i,j\}} = \left\{ p \in \Delta_M : halfSpace(i, j)(\ell_i - \ell_j)^\top p > 0 \right\}.$$

The function GETPOLYTOPE calculates the open polytope defined as the intersection of the above halfspaces. Then for all $i \in \mathcal{P}$ it checks if $\mathcal{C}_i$ intersects with the open polytope. If so, then $i$ will be an element of $\mathcal{P}(t)$. Similarly, for every $\{i, j\} \in \mathcal{N}$, it checks if $\mathcal{C}_i \cap \mathcal{C}_j$ intersects with the open polytope and puts the pair in $\mathcal{N}(t)$ if it does.

Note that it is not enough to compute $\mathcal{P}(t)$ and then drop from $\mathcal{N}$ those pairs $\{k, l\}$ where one of $k$ or $l$ is excluded from $\mathcal{P}(t)$: it is possible that the boundary $\mathcal{C}_k \cap \mathcal{C}_l$ between the cells of two actions $k, l \in \mathcal{P}(t)$ is included in the rejected region. For an illustration of cell decomposition and excluding cells, see Figure 1.

**Computational complexity** The computationally heavy parts of the algorithm are the initial calculation of the cell decomposition and the function GETPOLYTOPE. All of these require linear programming. In the preprocessing phase we need to solve $N + N^2$ linear programs to determine cells and neighboring pairs of cells. Then in every round, at most $N^2$ linear programs are needed. The algorithm can be sped up by "caching" previously solved linear programs.

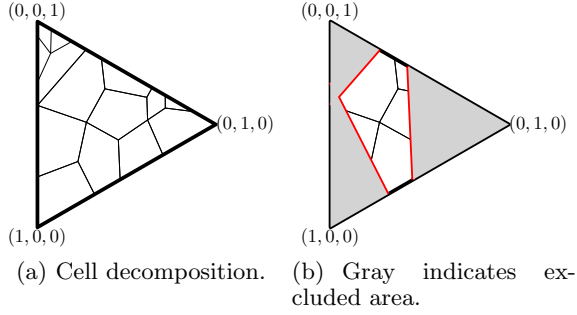

(a) Cell decomposition.

(b) Gray indicates excluded area.

*Figure 1.* An example of cell decomposition ($M = 3$).

## 4. Analysis of the algorithm

The first theorem in this section is an individual upper bound on the regret of CBP.

**Theorem 1.** *Let $(L, H)$ be an $N$ by $M$ partial-monitoring game. For a fixed opponent strategy $p^* \in \Delta_M$, let $\delta_i$ denote the difference between the expected loss of action $i$ and an optimal action. For any time horizon $T$, algorithm CBP with parameters $\alpha > 1$, $\nu_k = W_k^{2/3}$, $f(t) = \alpha^{1/3} t^{2/3} \log^{1/3} t$ has expected regret*

$$
\mathbb{E}[R_T] \leq \sum_{\{i,j\} \in \mathcal{N}} 2|V_{i,j}| \left( 1 + \frac{1}{2\alpha - 2} \right) + \sum_{k=1}^{N} \delta_k
$$
$$
+ \sum_{\substack{k=1 \\ \delta_k > 0}}^{N} 4W_k^2 \frac{d_k^2}{\delta_k} \alpha \log T
$$
$$
+ \sum_{k \in \mathcal{V} \setminus N^+} \delta_k \min \left( 4W_k^2 \frac{d_{l(k)}^2}{\delta_{l(k)}^2} \alpha \log T, \right.
$$
$$
\left. \alpha^{1/3} W_k^{2/3} T^{2/3} \log^{1/3} T \right)
$$
$$
+ \sum_{k \in \mathcal{V} \setminus N^+} \delta_k \alpha^{1/3} W_k^{2/3} T^{2/3} \log^{1/3} T
$$
$$
+ 2d_k \alpha^{1/3} W^{2/3} T^{2/3} \log^{1/3} T,
$$

*where $W = \max_{k \in \underline{N}} W_k$, $\mathcal{V} = \cup_{\{i,j\} \in \mathcal{N}} V_{i,j}$, $N^+ = \cup_{\{i,j\} \in \mathcal{N}} N_{i,j}^+$, and $d_1, \ldots, d_N$ are game-dependent constants.*

The proof is omitted for lack of space.[4] Here we give a short explanation of the different terms in the bound. The first term corresponds to the confidence interval failure event. The second term comes from the initialization phase of the algorithm. The remaining four terms come from categorizing the choices of the algorithm by two criteria: (1) Would $I_t$ be different if $\mathcal{R}(t)$ was defined as $\mathcal{R}(t) = \underline{N}$? (2) Is $I_t \in \mathcal{P}(t) \cup N_t^+$? These two binary events lead to four different cases in the proof, resulting in the last four terms of the bound.

[4]For complete proofs we refer the reader to the supplementary material.

An implication of Theorem 1 is an upper bound on the individual regret of locally observable games:

**Corollary 1.** *If $G$ is locally observable then*

$$
\mathbb{E}[R_T] \leq \sum_{\{i,j\} \in \mathcal{N}} 2|V_{i,j}| \left( 1 + \frac{1}{2\alpha - 2} \right)
$$
$$
+ \sum_{k=1}^{N} \delta_k + 4W_k^2 \frac{d_k^2}{\delta_k} \alpha \log T .
$$

*Proof.* If a game is locally observable then $\mathcal{V} \setminus N^+ = \emptyset$, leaving the last two sums of the statement of Theorem 1 zero. □

The following corollary is an upper bound on the minimax regret of any globally observable game.

**Corollary 2.** *Let $G$ be a globally observable game. Then there exists a constant $c$ such that the expected regret can be upper bounded independently of the choice of $p^*$ as*

$$
\mathbb{E}[R_T] \leq cT^{2/3} \log^{1/3} T .
$$

The following theorem is an upper bound on the minimax regret of any globally observable game against "benign" opponents. To state the theorem, we need a new definition. Let $A$ be some subset of actions in $G$. We call $A$ a *point-local game* in $G$ if $\bigcap_{i \in A} \mathcal{C}_i \neq \emptyset$.

**Theorem 2.** *Let $G$ be a globally observable game. Let $\Delta' \subseteq \Delta_M$ be some subset of the probability simplex such that its topological closure $\overline{\Delta'}$ has $\overline{\Delta'} \cap \mathcal{C}_i \cap \mathcal{C}_j = \emptyset$ for every $\{i, j\} \in \mathcal{N} \setminus \mathcal{L}$. Then there exists a constant $c$ such that for every $p^* \in \Delta'$, algorithm CBP with parameters $\alpha > 1$, $\nu_k = W_k^{2/3}$, $f(t) = \alpha^{1/3} t^{2/3} \log^{1/3} t$ achieves*

$$
\mathbb{E}[R_T] \leq cd_{pmax} \sqrt{bT \log T} ,
$$

*where $b$ is the size of the largest point-local game, and $d_{pmax}$ is a game-dependent constant.*

In a nutshell, the proof revisits the four cases of the proof of Theorem 1, and shows that the terms which would yield $T^{2/3}$ upper bound can be non-zero only for a limited number of time steps.

*Remark* 1. Note that the above theorem implies that CBP does not need to have any prior knowledge about $\Delta'$ to achieve $\sqrt{T}$ regret. This is why we say our algorithm is "adaptive".

An immediate implication of Theorem 2 is the following minimax bound for locally observable games:

**Corollary 3.** *Let $G$ be a locally observable finite partial monitoring game. Then there exists a constant $c$ such that for every $p \in \Delta_M$,*

$$\mathbb{E}[R_T] \le c\sqrt{T \log T}.$$

*Remark* 2. The upper bounds in Corollaries 2 and 3 both have matching lower bounds up to logarithmic factors (Bartók et al., 2011), proving that CBP achieves near optimal regret in both locally observable and non-locally observable games.

## 5. Experiments

We demonstrate the results of the previous sections using instances of Dynamic Pricing, as well as a locally observable game. We compare the results of CBP to two other algorithms: BALATON (Bartók et al., 2011) which is, as mentioned earlier in the paper, the first algorithm that achieves $\widetilde{O}(\sqrt{T})$ minimax regret for all locally observable finite stochastic partial-monitoring games; and FeedExp3 (Piccolboni & Schindelhauer, 2001), which achieves $O(T^{2/3})$ minimax regret on all non-hopeless finite partial-monitoring games, even against adversarial opponents.

### 5.1. A locally observable game

The game we use to compare CBP and BALATON has 3 actions and 3 outcomes. The game is described with the loss and feedback matrices:

$$L = \begin{pmatrix} 1 & 1 & 0 \\ 0 & 1 & 1 \\ 1 & 0 & 1 \end{pmatrix}; \qquad H = \begin{pmatrix} a & b & b \\ b & a & b \\ b & b & a \end{pmatrix}.$$

We ran the algorithms 10 times for 15 different stochastic strategies. We averaged the results for each strategy and then took pointwise maximum over the 15 strategies. Figure 2(a) shows the empirical minimax regret calculated the way described above. In addition, Figure 2(b) shows the regret of the algorithms against one of the opponents, averaged over 100 runs. The results indicate that CBP outperforms both FeedExp and BALATON. We also observe that, although the asymptotic performace of BALATON is proven to be better than that of FeedExp, a larger constant factor makes BALATON lose against FeedExp even at time step ten million.

### 5.2. Dynamic Pricing

In Dynamic Pricing, at every time step a seller (player) sets a price for his product while a buyer (opponent) secretly sets a maximum price he is willing to pay. The feedback for the seller is "buy" or "no-buy", while his loss is either a preset constant (no-buy) or the difference between the prices (buy). The finite version of the game can be described with the following matrices:

$$L = \begin{pmatrix} 0 & 1 & \cdots & N-1 \\ c & 0 & \cdots & N-2 \\ \vdots & \ddots & \ddots & \vdots \\ c & \cdots & c & 0 \end{pmatrix} \quad H = \begin{pmatrix} y & y & \cdots & y \\ n & y & \cdots & y \\ \vdots & \ddots & \ddots & \vdots \\ n & \cdots & n & y \end{pmatrix}$$

This game is not locally observable and thus it is "hard" (Bartók et al., 2011). Simple linear algebra gives that the locally observable action pairs are the "consecutive" actions ($\mathcal{L} = \{\{i, i+1\} : i \in \underline{N-1}\}$), while quite surprisingly, all action pairs are neighbors.

We compare CBP with FeedExp on Dynamic Pricing with $N = M = 5$ and $c = 2$. Since BALATON is undefined on not locally observable games, we can not include it in the comparison. To demonstrate the adaptiveness of CBP, we use two sets of opponent strategies. The "benign" setting is a set of opponents which are far away from "dangerous" regions, that is, from boundaries between cells of non-locally observable neighboring action pairs. The "harsh" settings, however, include opponent strategies that are close or on the boundary between two such actions. For each setting we maximize over 15 strategies and average over 10 runs. We also compare the individual regret of the two algorithms against one benign and one harsh strategy. We averaged over 100 runs and plotted the 90 percent confidence intervals.

The results (shown in Figures 3 and 4) indicate that CBP has a significant advantage over FeedExp on benign settings. Nevertheless, for the harsh settings FeedExp slightly outperforms CBP, which we think is a reasonable price to pay for the benefit of adaptivity.

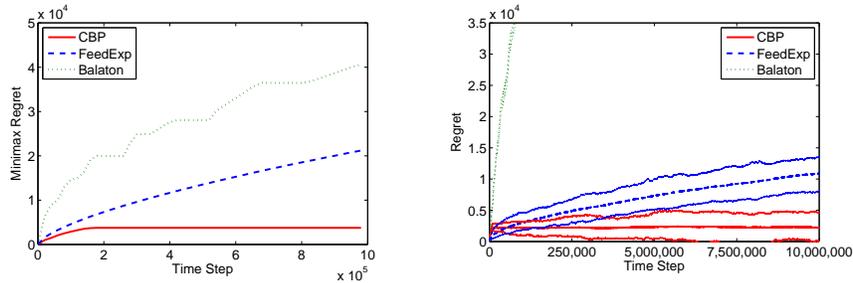

(a) Pointwise maximum over 15 settings.    (b) Regret against one opponent strategy.

*Figure 2.* Comparing CBP with BALATON and FeedExp on the easy game

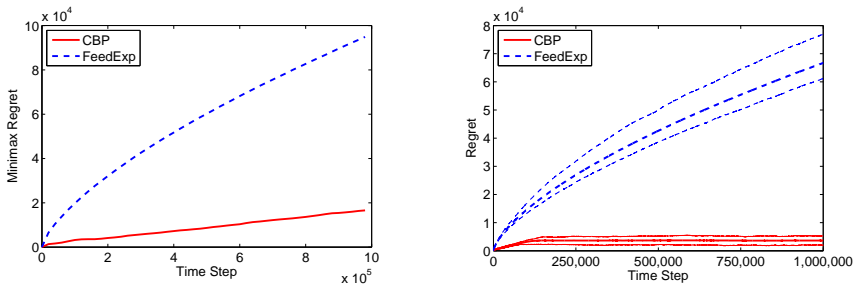

(a) Pointwise maximum over 15 settings.    (b) Regret against one opponent strategy.

*Figure 3.* Comparing CBP and FeedExp on "benign" setting of the Dynamic Pricing game.

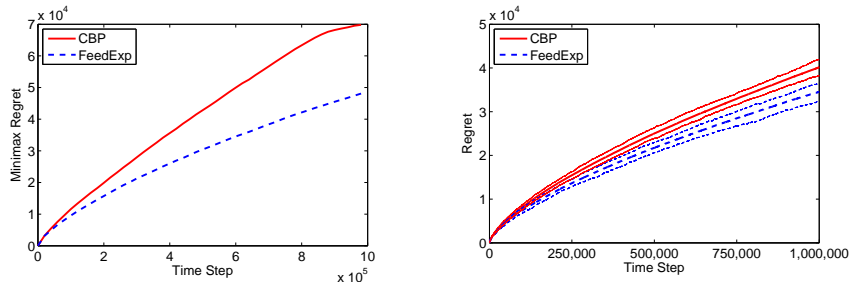

(a) Pointwise maximum over 15 settings.    (b) Regret against one opponent strategy.

*Figure 4.* Comparing CBP and FeedExp on "harsh" setting of the Dynamic Pricing game.